# Convolutional Long Short-Term Memory Neural Networks Based Numerical Simulation of Flow Field

Chang Liu

School of Computer Science and Engineering, University of Electronic Science and Technology of China, Chengdu 611731, China

**Abstract.** Computational Fluid Dynamics (CFD) is the main approach to analyzing flow field. However, the convergence and accuracy depend largely on mathematical models of flow, numerical methods, and time consumption. Deep learning-based analysis of flow filed provides an alternative. For the task of flow field prediction, an improved Convolutional Long Short-Term Memory (ConvLSTM) Neural Network is proposed as the baseline network in consideration of the temporal and spatial characteristics of flow field. Combining dynamic mesh technology and User-Defined Function (UDF), numerical simulations of flow around a circular cylinder were conducted. Flow field snapshots were used to sample data from the cylinder's wake region at different time instants, constructing a flow field dataset with sufficient volume and rich flow state variations. Residual networks and attention mechanisms are combined with the standard ConvLSTM model. Compared with the standard ConvLSTM model, the results demonstrate that the improved ConvLSTM model can extract more temporal and spatial features while having fewer parameters and shorter training time.

**Keywords:** Convolutional Neural Networks, Long Short-Term Memory, Numerical simulation, Flow field, Attention mechanism.

## 1       Introduction

Fluid mechanics, as a fundamental theoretical science, has been widely applied in civil engineering, aerospace, medical science, and energy with the rapid development of modern technology. Research methods in fluid mechanics can be broadly categorized into three main approaches: theoretical studies, experimental research, and numerical computation. With advancements in computing technology, computational power has grown exponentially. Numerical simulations, due to their independence from costly wind tunnel constructions and shorter experimental cycles, have become a widely adopted approach in fluid mechanics research. Today, Computational Fluid Dynamics (CFD) stands as the dominant methodology in the field. However, numerical simulations still face limitations, including high dependency on mathematical models, stringent requirements for numerical methods, significant time consumption, and computational uncertainties.

The rise of deep learning has introduced new possibilities for addressing these challenges. Currently, integrating deep learning with fluid mechanics and extracting empirical knowledge from vast datasets has emerged as a trending topic, often referred to as "intelligent fluid dynamics." In scenarios requiring rapid iterations such as industrial design optimization, traditional CFD methods demand substantial time and effort. Therefore, deep learning-based flow field simulation methods hold significant research value, as they can drastically reduce validation time in industrial design, accelerate development cycles, and enhance efficiency, particularly in computationally intensive simulation scenarios. Lee and You [1] pioneered the integration of CFD with GANs (Generative Adversarial Networks). They conducted numerical simulations under varying Reynolds numbers and time steps, using the resulting flow field variables as training datasets. By employing a GAN framework, they successfully predicted unsteady laminar vortex shedding around a cylinder. The study demonstrated that the predicted flow fields exhibited both qualitative and quantitative agreement with CFD-computed results, validating that deep learning techniques can effectively replace Navier-Stokes equations for laminar flow prediction. Lee and You [2] employed both conservative and non-conservative GAN architectures along with Convolutional Neural Networks (CNNs) to predict unsteady cylinder wake flows. Their innovative approach introduced a physics-informed loss function that explicitly incorporated mass and momentum conservation constraints into the deep learning framework. For long-term predictions, they recursively fed predicted flow fields back as input data. The study demonstrated that GANs with physics-informed loss functions and multi-scale CNNs exhibited superior predictive capabilities compared to conventional CNNs without physical constraints. In architectural applications of fluid mechanics, Kastner [3] developed a GAN-based approximate CFD surrogate model capable of processing arbitrary urban morphologies and building massing models; Rodriguez [4] introduced a generative machine learning model for predicting airflow in architectural spaces; Hu et al. [5] utilized sensor-measured velocity data as input and CFD-simulated airflow distribution data as training/validation sets to develop a CWGAN network, which can predict unsteady flow fields in three-dimensional building cluster models and rapidly estimate airflow distributions in urban models. Cori et al. [6] extended recurrent neural networks and pioneered the concept of Graph Neural Networks (GNNs), which can be applied to various graph data structures including directed graphs, undirected graphs, labeled graphs, and cyclic graphs. Their work successfully extended deep neural networks, originally designed for Euclidean data, to non-Euclidean domains, inspiring numerous researchers to explore graph neural networks. For graph-structured data, Chami et al. [7] introduced a unified framework to compare different machine learning models, and proposed a generalized framework encompassing shallow graph embedding methods, graph autoencoders, graph regularization approaches, and graph neural networks. Kipf and Welling [8] drew inspiration from graph operations in neural networks and semi-supervised learning on graphs to develop a novel semi-supervised classification method capable of encoding both graph structure and node features. Their results demonstrated that this approach could significantly improve computational efficiency. Bhatnagar et al. [9] proposed an encoder-decoder based surrogate model for predicting velocity and pressure fields. Their model utilized flow field data around various airfoils under different operating conditions, computed using Reynolds-Averaged Navier-Stokes (RANS) equations. The network inputs included Reynolds number, angle of attack, and airfoil geometry represented as signed distance functions, while outputs were velocity and pressure field distributions. Although this model provided new insights for design optimization, its generalization capability remained limited since only three airfoil shapes were used for training. Ribeiro et al. [10] developed the model (DeepCFD model) based on CNN architecture, which effectively approximated solutions for non-uniform steady laminar flow problems. Building upon DeepCFD, Abucide et al. [11] solved steady turbulent Navier-Stokes equations with variable input velocities. Portal and Abucide [12] constructed a U-Net structured CNN model that took signed distance functions, flow domain channels, and N samples as inputs, with CFD-computed data as outputs for training. This model successfully predicted streamwise and vertical velocities along with pressure fields downstream of cylinders. They later improved this architecture by proposing an autoencoder-based variant. Portal-Porra et al. [13] advanced this line of research by developing a CNN capable of predicting turbulence for different geometries and scales through approximating realizable $k$-$\epsilon$ two-equation turbulence models based on RANS equations. Wiewel et al. [14] employed a CNN to encode multi-step simulation fields, particularly pressure fields, into a reduced-order latent representation. This was then combined with a Long Short-Term Memory Network (LSTM) to perform single-step or multistep predictions of flow fields in the latent space. Their model demonstrated over 150 times faster computation compared to conventional pressure solvers. The study confirmed that the combination of LSTM and CNN is well-suited for flow field prediction tasks. Ko et al. [15] employed a Stream-wise Bidirectional Long Short-Term Memory neural network (SB-LSTM) to predict intra-channel flow. Wu [16] adopted a Reduced Order Modeling (ROM) approach based on Gated Recurrent Unit (GRU) networks for aerodynamic force identification and prediction during large-amplitude pitching oscillations of airfoils.

Although deep learning has achieved rapid development across multiple engineering and scientific disciplines, research on deep learning-based numerical simulation of flow fields remains in its nascent stage overall, with the powerful capabilities of deep learning yet to be fully realized. This paper explores deep learning-based numerical simulation of flow fields and proposes an improved ConvLSTM model that integrates attention mechanisms, 3D convolutional networks, and residual modules. By combining this module with a Convolutional Long Short-Term Memory Network, flow field prediction is achieved. The main contributions include

(1) The simulation of flow around a single cylinder was conducted using dynamic mesh technology and User-Defined Functions (UDF). The computational results were compared with published literature data, verifying the validity of the numerical simulation methodology employed in this study.

(2) Residual networks and attention mechanisms are combined with the standard ConvLSTM model. Compared to the standard ConvLSTM, the improved model extracts richer spatiotemporal features with lower computational cost and faster training.

## 2   Construction of Flow Field Dataset Based on Numerical Simulation

Deep learning requires substantial amounts of data, but obtaining flow field data through experimental methods faces challenges such as high costs and experimental uncertainties. Numerical simulation provides an effective alternative for acquiring flow field data. In this study, CFD (Computational Fluid Dynamics) methods were employed to construct a flow field dataset. Numerical simulations of flow around a circular cylinder were performed to investigate the generation of vortex streets in the wake region behind the cylinder, and the reliability of the data was validated.

### 2.1   Governing Equations

The fundamental equations of fluid dynamics are a system of equations that describe fluid motion, primarily including the continuity equation, momentum equation, and energy equation. By solving these equations with appropriate boundary and initial conditions, key information such as the fluid's motion state, pressure distribution, and velocity field can be obtained, enabling the analysis and prediction of fluid behavior.

Continuity equation:

$$\frac{\partial \rho}{\partial t} + \frac{\partial(\rho u)}{\partial x} + \frac{\partial(\rho v)}{\partial y} + \frac{\partial(\rho w)}{\partial z} = 0 \tag{1}$$

where $u$, $v$, $w$ denote the velocities; $\rho$ stands for the fluid density.

Momentum equation:

$$\begin{cases} \rho \frac{du}{dt} = -\frac{\partial p}{\partial x} + \frac{\partial \tau_{xx}}{\partial x} + \frac{\partial \tau_{yx}}{\partial y} + \frac{\partial \tau_{zx}}{\partial z} + \rho f_x \\ \rho \frac{dv}{dt} = -\frac{\partial p}{\partial y} + \frac{\partial \tau_{xy}}{\partial x} + \frac{\partial \tau_{yy}}{\partial y} + \frac{\partial \tau_{zy}}{\partial z} + \rho f_y \\ \rho \frac{dw}{dt} = -\frac{\partial p}{\partial z} + \frac{\partial \tau_{xz}}{\partial x} + \frac{\partial \tau_{yz}}{\partial y} + \frac{\partial \tau_{zz}}{\partial z} + \rho f_z \end{cases} \tag{2}$$

where $p$ denotes the surface pressure; $f_x$, $f_y$, $f_z$ represent the components of the body force acting on the fluid element in the $x$, $y$, and $z$ directions, respectively; $\sigma_{ij}$ indicates the component of the stress tensor acting on the fluid element.

Energy equation:

$$\frac{\partial(\rho T)}{\partial t} = div(ruT) = div\left(\frac{K}{C_P} grad(T)\right) + S_T \tag{3}$$

where $T$ represents the thermodynamic temperature; $K$ stands for the heat transfer coefficient; $C_p$ indicates the specific heat capacity at constant pressure; $S_T$ symbolizes the viscous dissipation term.

## 2.2 Numerical Simulation

The flow around tandem circular cylinders are commonly encountered in engineering applications and natural phenomena, such as the wake flows behind structures like bridges, buildings, and pipelines. In this configuration, two cylinders are arranged in-line, generating periodic vortex street patterns due to their mutual interference. The wake flow field exhibits three classical patterns: the no-vortex mode, single-vortex mode, and double-vortex mode. Based on the Reynolds similarity principle, this study conducts numerical simulations of flow around tandem circular cylinders at a Reynolds number of 200. The geometric model and computational domain configuration are illustrated in Fig. 1, where $L_u$ represents the distance from the upstream cylinder center to the inlet boundary; $L$ denotes the center-to-center spacing between the upstream and downstream cylinders; $L_d$ indicates the distance from the downstream cylinder center to the outlet boundary; $W$ stands for the width of the computational domain.

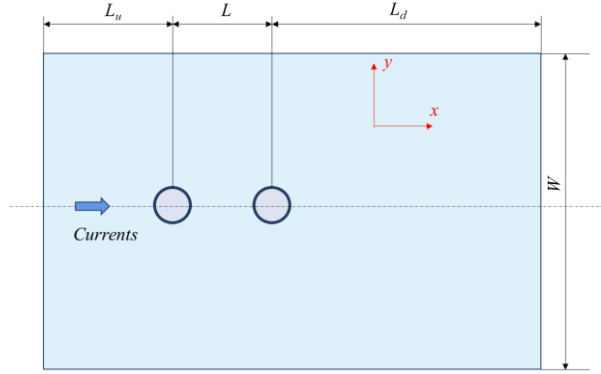

**Fig. 1.** Geometric model and computational domain.

This study employs unstructured grids, which offer superior flexibility in mesh generation. To optimize computational resources and enhance efficiency, an overset grid (chimera grid) approach is adopted, consisting of two distinct components: background grid and moving grid. Both grids are created using the ICEM CFD module in ANSYS. For background grid, it has coarser mesh resolution while covers the global computational domain and handles far-field flow conditions. For moving grid, it has finer mesh density concentrated near the central region and resolves critical flow features (e.g. boundary layers, vortex shedding).

To meet the requirements of low-Reynolds-number flow simulations, a grid growth rate of 1.1 was adopted. Computational results show that the first-layer grid height is $y^+ < 10$. At least 5 grid layers are arranged within the boundary layer. Quality verification confirms the computational grids exhibit excellent characteristics. The minimum grid quality reaches 0.908 across three different inter-cylinder spacing cases. All grids satisfy the precision requirements for numerical simulations. Boundary conditions are configured as follows: velocity inlet boundary condition with specified velocity $U$=0.3m/s; pressure outlet boundary condition; no-slip wall boundary condition;

medium density $\rho$=1.0kg/m3, dynamic viscosity $\mu = 1.5 \times 10^{-5}$kg/ (m·s). Solver settings include: laminar flow model; residual convergence criterion set to $1\times10^{-5}$; time step size with 0.001 s; 20,000 iterations. Monitoring parameters refer to lift coefficient ($C_L$) and drag coefficient ($C_D$) of both upstream and downstream cylinders. Simulation results regarding the velocity distribution around cylinders are shown as Fig. 2. As evident from the results, distinct dual-vortex structures are clearly observed, which demonstrates agreement with findings reported in the literature.

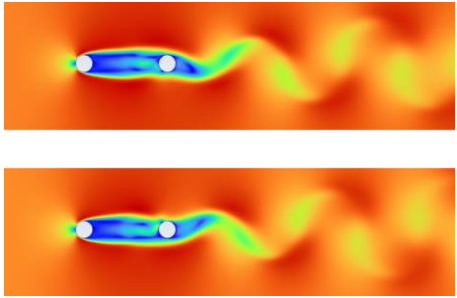

**Fig. 2.** Numerical simulation results by CFD.

### 2.3    Dataset

Based on the simulation results, dataset for the training and validation in deep learning is constructed. Although the numerical simulation used a time step of 0.001s, adopting this interval for dataset construction would result in negligible observable changes in the wake flow field behind the cylinder. To enable the neural network to learn more complex flow characteristics, the sampling interval is set as 0.02s. Additionally, to ensure dataset validity and authenticity, data from the initial transient phase of the simulation were excluded. The dataset was partitioned using a random seed generation method, with the following allocation: 70% of the data was assigned to the training set; 10% to the validation set; 20% to the test set.

## 3    ConvLSTM with Residual Module and Attention Mechanism

Flow field data represents a typical spatiotemporal sequence. The data at any given point in the flow field exhibits both temporal correlation with adjacent time steps and spatial correlation with neighboring points in the current flow field. The Convolutional Long Short-Term Memory (ConvLSTM) network demonstrates unique advantages for spatiotemporal sequence prediction tasks due to its distinctive recurrent architecture, which enables long-term memory capability in spatiotemporal domains and facilitates in-depth extraction of latent features from spatiotemporal data. Therefore, this study employs ConvLSTM as the foundation for constructing the flow field prediction model. Building upon the standard ConvLSTM architecture, the model's predictive accuracy is further enhanced by incorporating three key components: attention mechanism, residual connections, and 3D convolutional networks.

### 3.1 Convolutional Long Short-Term Memory Neural Networks

The Convolutional Long Short-Term Memory network (ConvLSTM) is an advanced architecture developed by augmenting the traditional Long Short-Term Memory (LSTM) network. This hybrid model combines the distinctive strengths of two fundamental neural network paradigms, i.e., Convolutional Neural Networks (CNNs) and Long Short-Term Memory Networks. The ConvLSTM's key innovation lies in its replacement of standard matrix multiplications in LSTM gates with convolutional operations.

$$\begin{cases} g_t = tanh(W_{xg} * X_t + W_{hg} * H_{t-1} + b_g) \\ i_t = \sigma(W_{xi} * X_t + W_{hi} * H_{t-1} + W_{ci} \odot C_{t-1} + b_i) \\ f_t = \sigma(W_{xf} * X_t + W_{hf} * H_{t-1} + W_{cf} \odot C_{t-1} + b_f) \\ C_t = f_t \odot C_{t-1} + i_t \odot g_t \\ o_t = \sigma(W_{xo} * X_t + W_{ho} * H_{t-1} + W_{co} \odot C_t + b_o) \\ H_t = o_t \odot tanh(C_t) \end{cases} \quad (4)$$

where $i_t$ denotes the input gate; $f_t$ the forget gate; $o_t$ the output gate; $g_t$ the update gate. $X_t$ is the input at time t; $H_{t-1}$ is the hidden state at time $t-1$; $\odot$ denotes the Hadamard product while $*$ denotes the convolution operation.

### 3.2 Residual Networks

To more fully extract the spatial features of data, scholars often construct networks with more layers. However, many experimental results show that deeper networks do not always perform better. Excessively deep networks may lead to higher training errors, and training such networks becomes increasingly difficult. To address this issue, Residual Networks (ResNet) was proposed [17]. ResNet mitigates the problems of vanishing and exploding gradients, improves network performance and generalization capabilities, accelerates training convergence, and thus enables the training of very deep networks. A representative ResNet unit is shown as Fig. 3.

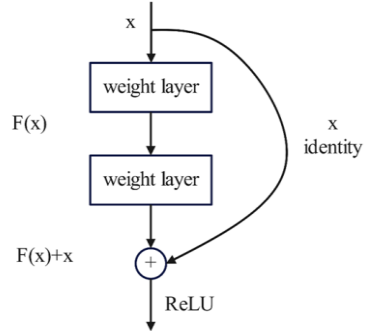

**Fig. 3.** ResNet unit.

The data transmission method of the residual module can be described as

$$\begin{cases} y_l = F(x_l, W_l) + x_l \\ x_{l+1} = f(y_l) \end{cases} \quad (5)$$

where $f(\cdot)$ denotes ReLU activation function; $x_l$ and $x_{l+1}$ denote the input and output of the ResNet unit. Based on (5), the learning features from shallow layer 1 to deep layer $L$ can be obtained as

$$x_L = x_1 + \sum_{i=1}^{L-1} F(x_i, W_i) \quad (6)$$

### 3.3  Channel Attention Mechanism

The Channel Attention Mechanism (CAM) is a category of attention mechanisms that fully leverages the information from each channel to enhance the model's representational capacity for input data. Among these, the most representative is the Squeeze-and-Excitation Network (SENet).

A representative SENet is shown as Fig. 4. The squeeze operation performs global average pooling on the input feature map, compressing the features of each channel to produce a compact global descriptor vector. The excitation operation processes the obtained weights through fully connected layers, calculating the importance of each channel. The higher the weight, the greater the attention assigned to that channel. The scale operation adjusts or normalizes the weighted features to ensure the output features maintain an appropriate range and magnitude.

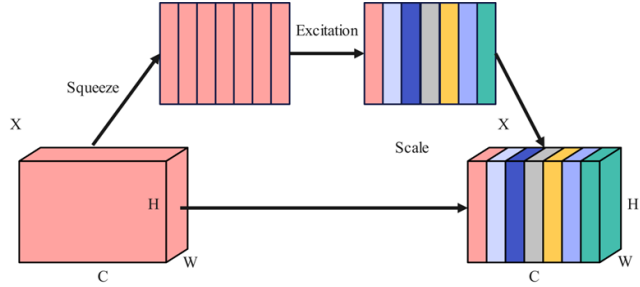

**Fig. 4.** SENet unit.

### 3.4  ConvLSTM with ReNet and SENet

Combining the attention mechanism, residual network, and 3D convolution, an improved ConvLSTM architecture is illustrated in Fig. 5. This network primarily consists of an SENet-ResNet module, ConvLSTM module, and a Conv3D module. Among them, the main role of the Conv3D module is not feature extraction but rather transforming the dimensionality of the feature data. The SENet-ResNet module is mainly composed of a 3D residual network and a channel attention mechanism, with its primary function being the extraction of spatiotemporal features from the flow field. Unlike pure time series, flow field data is a typical spatiotemporal sequence. The data at a given point in the flow field is not only temporally correlated with the flow field at adjacent time steps but also spatially correlated with surrounding points in the flow field at the current moment. To better extract the spatiotemporal features from the flow field, in the study ReNet and SENet modules are combined with ConvLSTM.

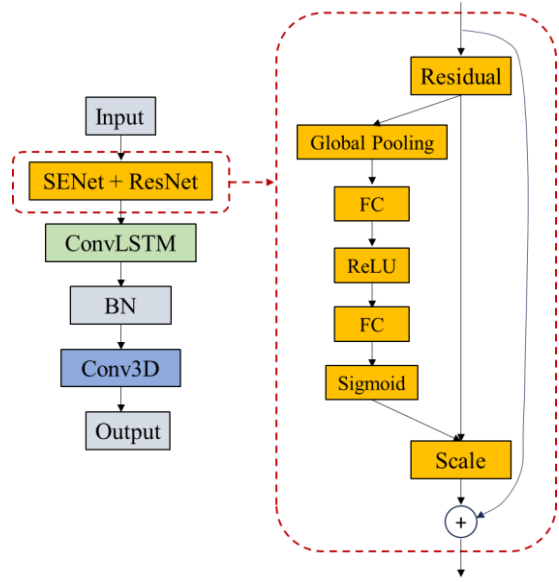

**Fig. 5.** Structure of ConvLSTM with ReNet and SENet.

## 4    Prediction of Flow Field by ConvLSTM

The experiment was conducted in a Python, Keras, and TensorFlow environment, with the respective versions being Python 3.9.0, Keras 2.11.0, and TensorFlow 2.11.0. Additionally, the computer operating system was Windows 10, with the following hardware configuration: Processor: Intel Core i5-10600KF (6 cores, 12 threads); Graphics card: NVIDIA GeForce RTX 3080 (10GB); CUDA version: 11.2; cuDNN (GPU acceleration library) version: 8.0.4; RAM capacity: 32GB.

To validate the effectiveness of the deep learning-based flow field prediction model proposed in this study, a combined approach of visual assessment and objective metrics are employed. First, the predicted flow field is visually compared with the original flow field. Then, quantitative evaluations are conducted using the mean absolute error (MAE), mean squared error (MSE), and structural similarity index (SSIM).

$$MAE = \frac{1}{n}\sum_{i=1}^{n}\left|Y_i - \hat{Y}_i\right| \tag{7}$$

$$MSE = \frac{1}{n}\sum_{i=1}^{n}\left(Y_i - \hat{Y}_i\right)^2 \tag{8}$$

$$SSIM(x,y) = \frac{(2\mu_x\mu_y + c_1)(2\sigma_{xy} + c_2)}{(\mu_x^2 + \mu_y^2 + c_1)(\sigma_x^2 + \sigma_y^2 + c_2)} \tag{9}$$

where $Y_i$ denotes the reference value while $\hat{Y}_i$ denotes the predicted value; $\mu_i$ denotes the average value; $\sigma_i^2$ denotes the variance; $c_i$ is constant.

Using the improved ConvLSTM model, the flow field is predicted. Prediction results of frame 11 and frame 14 are given in Fig. 5. As can be seen, the vortex characteristic can be well predicted by the proposed ConvLSTM model.

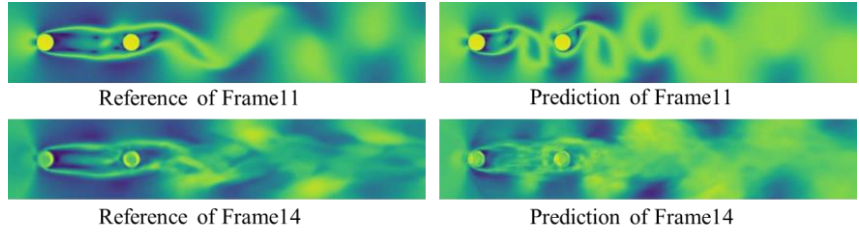

Reference of Frame11     Prediction of Frame11
Reference of Frame14     Prediction of Frame14

**Fig. 6.** Prediction of flow field by improved ConvLSTM.

Fig.7 shows the improvement when residual networks, attention mechanisms, and 3D convolutions are incorporated into ConvLSTM. As can be seen, the residual networks and attention mechanisms obviously benefit the ConvLSTM model.

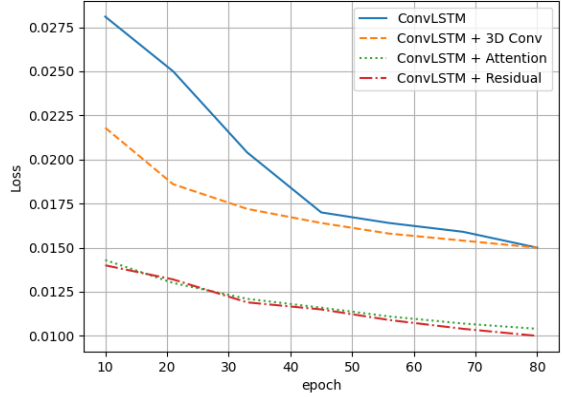

**Fig. 7.** Prediction of flow field by improved ConvLSTM.

Comparison is conducted to confirm the advantage of the proposed ConvLSTM over standard ConvLSTM. Table 1 lists the comparison results in terms of parameter capacity, training time, and three metrics including MAE, MSE and SSIM. As can be seen, the calculation burden is reduced significantly by using the proposed Con-vLSTM while the prediction accuracy is improved, which verifies the effectiveness and advantage of the proposed model.

Table 1. Comparison of standard ConvLSTM and improved ConvLSTM.

| Metric | Standard ConvLSTM | Improved ConvLSTM | Improvement |
|---|---|---|---|
| Total params | 495,457 | 310,385 | reduced by 37% |
| Trainable params | 495,265 | 310,161 | reduced by 37% |
| Training time (min) | 560 | 381 | reduced by 37% |
| MAE | 142 | 139 | reduced by 2% |
| MSE | 140 | 82 | reduced by 41% |
| SSIM | 0.71 | 0.84 | increased by 18% |

## 5  Conclusions

This study focuses on the prediction and reconstruction of flow fields by integrating deep learning with numerical simulations. Recognizing the critical role of datasets in training neural network models, datasets for tandem double-cylinder flow with varying inter-cylinder spacings are constructed. Based on ConvLSTM, a convolutional long short-term memory (ConvLSTM) flow field prediction model by incorporating channel attention mechanisms is developed, residual networks, and 3D convolutional modules, enabling the prediction of future-frame flow fields.

The ConvLSTM network, owing to its robust spatiotemporal feature extraction capability, enables the prediction of future flow field frames. By integrating the SE-Res3D module which combines channel attention mechanisms, residual networks, and 3D convolutions, the model achieves higher accuracy in flow field prediction. Compared to traditional ConvLSTM, the proposed architecture features fewer parameters and faster training speed. Nevertheless, it is noted that as the prediction time step increases, the prediction accuracy decreases accordingly, which is the limitation of the current approach. The author tried using dropout technique to deal with the issue but the improvement is trivial. Other measures will be studied in the next work.

The integration of fluid mechanics and deep learning remains in the exploratory research phase, and this study contributes some preliminary investigations. Regarding the application of deep learning to flow field prediction and reconstruction, several key challenges require further resolution. For example, while this work focuses on 2D planar flow field prediction, real-world engineering applications involve 3D flow fields. Subsequent research should explore 3D flow field prediction. Moreover, the scalability of the proposed model, especially in real-world, complex flow field scenarios will be studied. The comparison of the proposed model with other state-of-the-art models in terms of computational efficiency and prediction accuracy will be conducted in future work.

**Acknowledgments.** The authors appreciate the anonymous review for their help in improving the quality of the paper.

**Disclosure of Interests.** The authors have no competing interests to declare that are relevant to the content of this article.